\documentclass[]{CEURART-2/ceurart}

\usepackage{longtable,booktabs,array}
\usepackage{graphicx}
\usepackage{placeins}
\usepackage{url}
\usepackage{listings}
\makeatletter
\def\@copyrightLine{}
\makeatother
\ExplSyntaxOn
\cs_gset:Npn \__first_footerline: {}
\ExplSyntaxOff
\KOMAoptions{headsepline=false,footsepline=false}
\RenewDocumentCommand{\dashrule}{O{.4pt} m m}{}

\begin{document}

\conference{}

\title{DeepLens Diagnosis Agent: Agentic Workflow Design Lets a Small Reasoning Model Compete with Frontier LLMs}

\author[1]{Mahmood Bayeshi}
\author[1]{Veysel Kocaman}
\author[1]{Muhammed Ali Naqvi}
\author[1]{Yigit Gul}
\author[1]{David Talby}
\address[1]{John Snow Labs}

\begin{abstract}
Medical diagnosis is a multi-stage process: extract facts, consult knowledge, generate a differential analysis, and select the best diagnosis with explanations. Frontier LLMs are strong generalists, but single-shot prompting often yields brittle diagnostic reasoning. We present the DeepLens Diagnosis Agent, a five-stage \emph{harnessing} pipeline---the practice of combining model capabilities with disciplined process constraints---centered on a small medical reasoning model (JSL Medical Small 7B v2) \cite{jsl_medllm_docs} and retrieval-augmented generation (RAG) \cite{lewis2020rag}. The pipeline enforces structured clinical extraction, disciplined retrieval, constrained candidate generation, explicit evidence triangulation, and an auditable final decision. On a held-out 915-case diagnostic benchmark (DiagnosisArena) \cite{zhu2025diagnosisarena}, the agent achieved \textbf{60.14\%} top-1 diagnostic accuracy, the highest among small and medium-sized models. The same JSL Medical Small 7B v2 model without the agent workflow achieved \textbf{23.99\%}, a \textbf{+36-point gain} attributable to workflow design alone, despite the base model achieving 88.2\% on standard medical benchmarks---demonstrating that diagnostic reasoning under uncertainty requires more than medical knowledge recall. This accuracy gain is achieved at \textbf{\$0.0072 per case} (24K tokens per case on A100 infrastructure) with 24-second latency---35--45\% cheaper than frontier cloud models (Claude Sonnet 4.5: \$0.0110, Gemini 3.1 Pro: \$0.0128) while outperforming them by +9.70pp and +9.17pp respectively. Importantly, we demonstrate that harnessing and agentic diagnosis can improve frontier model failures, suggesting that workflow discipline is a complementary lever for all model scales. For diagnostic reasoning, workflow constraints and verification gates can outweigh parameter count or API costs.
Beyond aggregate accuracy, the pipeline produces structured intermediate artifacts that make each stage inspectable and facilitate error localization during failure analysis. These properties are essential for high-stakes deployment settings, where traceability, reproducibility, and reviewer-auditable evidence are necessary alongside strong benchmark performance.
\end{abstract}

\begin{keywords}
medical diagnosis \sep
agentic workflow \sep
retrieval-augmented generation \sep
differential diagnosis \sep
clinical reasoning
\end{keywords}

\maketitle

\section{Introduction}\label{introduction}

Medical diagnosis is fundamentally a workflow problem. Clinicians do not
produce diagnoses in one step: they structure patient facts, consult
knowledge, generate and refine a differential diagnosis, and finally
commit to a decision.

In contrast, many LLM baselines attempt implicit multi-step reasoning
within a single completion. This tends to blur \emph{extraction} (what
is stated) and \emph{inference} (what is implied), increasing error
propagation.

DiagnosisArena emphasizes this difficulty: long clinical narratives
require precise fact grounding (positives, negatives, timeline,
pathology) and multi-step inference to reach the correct diagnosis
\cite{zhu2025diagnosisarena}. More broadly, the field is trending toward structured
evaluation and disciplined medical reasoning workflows (e.g., medical
reasoning foundation models \cite{octomed2025} and expert review protocols)
\cite{kocaman2025clever}.

This paper advances the following thesis: a small reasoning model
embedded in a carefully engineered workflow can outperform larger models
when the system design is explicitly constrained. We call this approach
\emph{harnessing}---the practice of combining model capabilities with
disciplined process constraints to achieve reliable, auditable reasoning
at scale.

The motivation is practical as well as scientific. In real clinical
settings, diagnostic assistants are expected to be traceable, stable
across repeated runs, and easy to audit when outputs are disputed.
Single-pass free-form generation tends to optimize fluency, whereas
clinical deployment requires explicit control over evidence grounding,
error localization, and decision criteria. The workflow perspective
adopted in this paper addresses that requirement by making each stage
inspectable and by constraining how intermediate outputs can influence
the final diagnosis.

\begin{figure}[!htbp]
\centering
\includegraphics[width=1.0\linewidth]{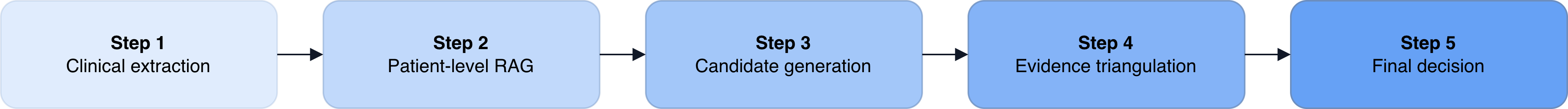}
\caption{Overview of the DeepLens Diagnosis Agent workflow and its
core reasoning stages.}
\label{fig:deeplens-workflow-overview}
\end{figure}

Accordingly, this paper focuses on system-level design choices rather
than model scaling alone (Figure~\ref{fig:deeplens-workflow-overview}
provides a high-level overview of the proposed pipeline). We show how
decomposition into extraction, supportive retrieval, constrained
candidate generation, and quote-anchored triangulation changes
diagnostic behavior in measurable ways. We then evaluate this design on DiagnosisArena and discuss why the
observed gains are consistent with workflow-level error reduction, not
just stronger parametric recall.

From an engineering perspective, this design can be interpreted as a
set of explicit reliability constraints applied to each reasoning stage.
Instead of optimizing for a single fluent output, the workflow optimizes
for traceable intermediate artifacts that can be validated and audited.
This shift is central to clinical decision support, where interpretability
and failure localization are often as important as benchmark-level
accuracy.

\section{Background and Related Work}\label{background-and-related-work}

Recent progress in medical LLMs shows strong gains on benchmarked QA,
but also highlights a persistent gap between benchmark accuracy and
reliable clinical reasoning under uncertainty. Early evidence that
general-purpose LLMs encode clinically useful knowledge came from
MultiMedQA/Med-PaLM style evaluations \cite{singhal2023medpalm}, and later work improved
specialist-grade performance via better medical alignment and prompting
strategies \cite{tu2025expertlevel}, \cite{nori2023medprompt}. In parallel, benchmark design evolved
toward harder diagnostic settings that require structured synthesis over
long clinical narratives instead of short fact recall \cite{zhu2025diagnosisarena}.

For this paper, the key background assumption is that diagnosis quality
depends not only on model knowledge, but also on reasoning-process
control. Chain-of-thought prompting showed that explicit intermediate
reasoning can improve task performance \cite{wei2022cot}, \cite{wang2022selfconsistency}, while ReAct-style
formulations demonstrated the value of combining reasoning with external
evidence access \cite{yao2022react}. In medical tasks, these findings imply that
architecture-level constraints (fact extraction, grounded comparison,
and explicit tie-breaks) can matter as much as raw parameter count.

Foundational medical LLM studies established strong performance on
standardized QA and licensing style benchmarks, but also documented
residual issues in calibration, omissions, and potential harm in
free-text outputs \cite{singhal2023medpalm}, \cite{tu2025expertlevel}. Recent frameworks such as CLEVER
push evaluation beyond label accuracy toward expert-reviewed quality
dimensions (clinical relevance, factuality, and justification fidelity),
which is especially important for decision-support claims \cite{kocaman2025clever}.

DiagnosisArena explicitly targets difficult diagnostic reasoning with
long-form case narratives and broad specialty coverage \cite{zhu2025diagnosisarena}. Compared
with multiple-choice medical QA, this benchmark exposes
workflow-sensitive failure modes: missing a key negative, mishandling
timeline constraints, or overfitting to a salient but nonspecific clue.

Retrieval-augmented generation provides a principled mechanism to inject
updatable external knowledge at inference time \cite{lewis2020rag}, \cite{mialon2023augmented}, and agentic
reasoning paradigms such as ReAct motivate separating planning, evidence
access, and decision actions \cite{yao2022react}, \cite{schick2023toolformer}. In healthcare, this decomposition
improves auditability because reviewers can inspect intermediate
artifacts rather than only a final narrative.

A key line of work asks whether strong prompting can close the gap with
domain-specialized models. Medprompt-style results suggest that
generalist models can achieve competitive medical benchmark performance
under carefully designed prompting \cite{nori2023medprompt}. However, for high-stakes
diagnosis, the literature still supports adding structured verification
gates and human expert review protocols, not relying on prompting alone
\cite{kocaman2025clever}.

Together, these studies motivate the approach in this paper: treat
diagnosis as a constrained, auditable workflow that combines structured
clinical extraction, supportive retrieval, candidate-level comparison,
and explicit final-decision guardrails.

DiagnosisArena is designed to stress diagnostic reasoning by requiring
models to interpret long clinical narratives and produce grounded
diagnoses \cite{zhu2025diagnosisarena}. Unlike multiple-choice medical QA benchmarks, it
requires accurate extraction of key positives and negatives, temporal
reasoning with timeline consistency, pathology and diagnostic test
interpretation when present, and differential diagnosis search under
uncertainty. The failure modes are also workflow-shaped: missing a single
negative, anchoring on a salient but non-specific clue, or inventing a
symptom can derail the final diagnosis. That makes the benchmark
appropriate for evaluating \emph{agentic decomposition} rather than only
raw knowledge recall, and motivates the pipeline-based approach
described in the following sections.

\FloatBarrier

\section{Methods: Five-Stage Diagnosis
Pipeline}\label{methods-five-stage-diagnosis-pipeline}

\begin{figure}[!htbp]
\centering
\includegraphics[width=0.88\linewidth]{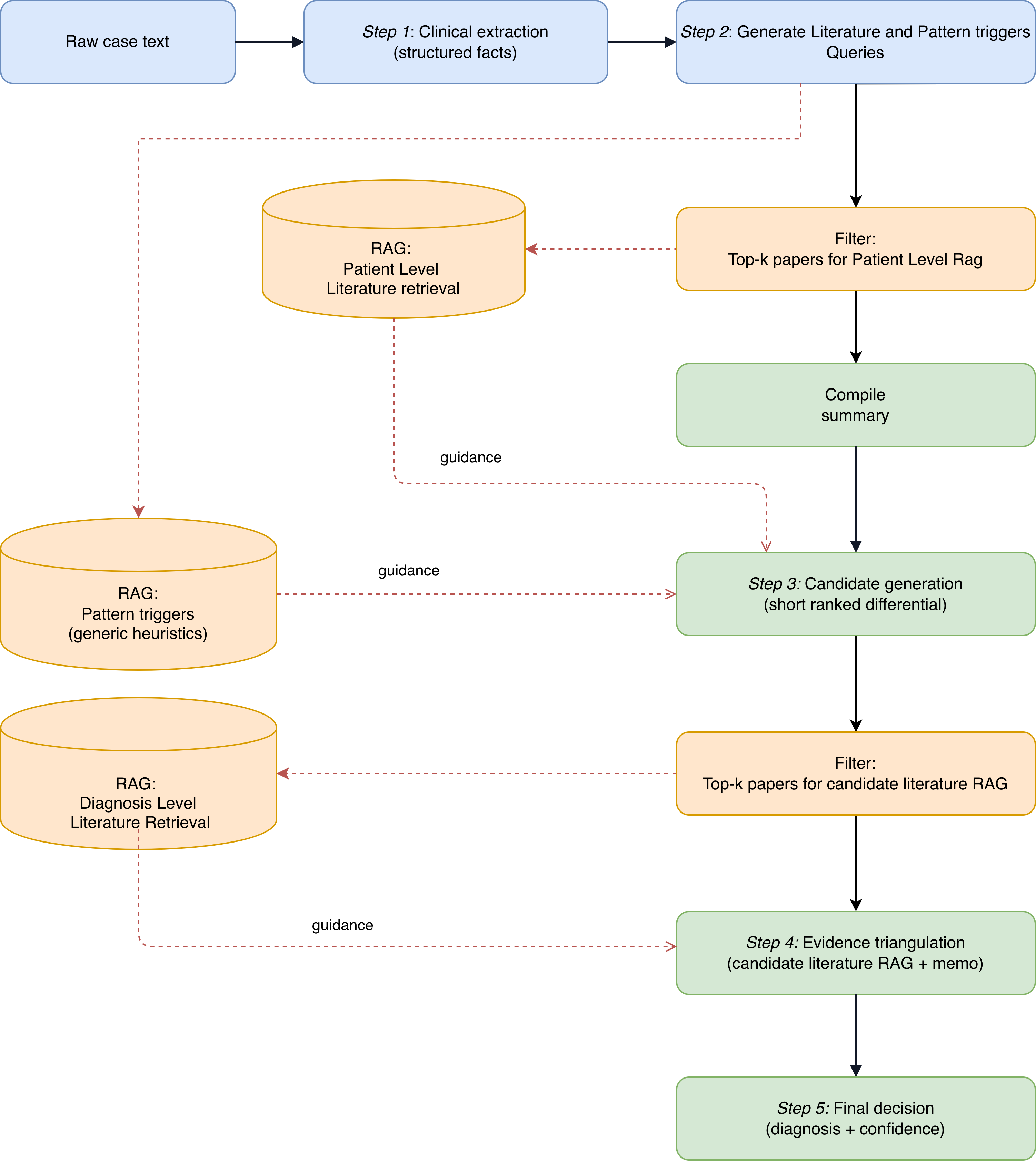}
\caption{Method pipeline used in this study for structured extraction,
retrieval support, candidate generation, and final diagnostic
selection.}
\label{fig:methods-pipeline}
\end{figure}

This section describes the five sequential stages of the Diagnosis Agent
pipeline. Figure~\ref{fig:methods-pipeline} illustrates the detailed
dataflow across stages.

\FloatBarrier

\subsection{Base Model: JSL Medical Small}

The Diagnosis Agent centers on JSL Medical Small, a 7B-parameter
medical language model fine-tuned for clinical reasoning tasks
\cite{jsl_medllm_docs}. On standard medical knowledge benchmarks, this model
demonstrates strong performance: 89.4\% on MedQA, 89\% on Anatomy,
95\% on College Biology, and 88.2\% average across nine standard
benchmarks \cite{jsl_medllm_docs}. This performance positions it as
competitive on narrow fact-recall tasks commonly used to evaluate
medical LLMs.

However, standard benchmarks do not measure diagnostic reasoning under
uncertainty. When deployed as a single-shot baseline on
DiagnosisArena---a benchmark emphasizing long-form case interpretation,
evidence synthesis, and differential diagnosis---JSL Medical Small
achieves only 23.99\% accuracy, a gap of approximately 64 percentage
points. This gap is not unique to this model: a larger 32B-parameter
variant from the same model family, which averages 85.0\% on standard
benchmarks, drops to 22.19\% on DiagnosisArena. The dramatic
performance cliff highlights a fundamental distinction: strong
performance on medical knowledge recall does not translate to
diagnostic workflow execution.

This observation motivates the core design: rather than scaling model
parameters, we embed the 7B model in a constrained pipeline that
enforces structured reasoning, explicit evidence grounding, and
decision verification. As demonstrated in Section~\ref{sec:results},
this approach recovers diagnostic accuracy to 60.14\%, a 36-point
improvement from the vanilla baseline and the highest accuracy among
smaller models---achieved through workflow discipline rather than
parameter scaling alone \cite{kocaman2025clever}.

\subsection{Clinical Extraction and Fact Lock-In}

The first step converts raw case text into a structured representation
that downstream steps treat as \emph{the only source of patient truth}.
This serves as a primary control against hallucination.

The extraction step produces a compact,
fixed set of fields that the rest of the pipeline treats as the
\emph{only} patient truth: \texttt{main\_case\_text} (core narrative),
\texttt{timeline}, \texttt{demographics}, \texttt{positives},
\texttt{negatives}, \texttt{diagnostics} (one test per item), and
\texttt{other\_information} (high-signal context, explicitly not a
differential).

In practice, this step enforces strict
machine-parseable formatting, extraction-only behavior, diagnostics
hygiene (atomic tests, preserved units, deduplication, specificity
ordering), anti-leakage constraints on \texttt{other\_information}, and
compact source-of-truth fields to preserve stability across downstream
steps.

The output is validated and made tolerant
to small-model shape mistakes (e.g., list-like fields returned as
strings are coerced into lists). The intent is to keep the workflow on
rails: downstream steps always receive the same \emph{kind} of object
(lists stay lists, required fields stay present), which reduces silent
failure modes.

As a result, the downstream model becomes a \emph{reasoner
over a fact table} rather than a generator operating on unstructured
prose.

\subsection{Patient-Level RAG and Pattern Triggers}

This stage builds two complementary supports for diagnosis: patient-level
RAG to retrieve relevant medical literature for the overall case \cite{lewis2020rag}
and pattern triggers to retrieve reusable non-disease-specific
diagnostic heuristics.

\subsubsection{Patient-Level RAG in a Supportive Role}

This stage retrieves broad medical literature to support the overall
case understanding. The agent generates one short retrieval query from the
extracted fact table, querying across Semantic Scholar and the John Snow
Labs curated medical knowledge base (200M+ scholarly records) using a
hybrid search strategy. Query-generation is intentionally strict: no
PHI-like identifiers (explicit ages, years, locations, names, or
institutions), preference for discriminators over generic symptoms, and
no citations/IDs/URLs or paper-numbering artifacts.

To protect privacy, the query is rejected if it
includes explicit ages (e.g., 50 years old) or 4-digit years. The
query is also kept intentionally short so retrieval stays focused on
discriminators rather than copying narrative details.

If LLM query generation fails, a heuristic
query builder composes an embedding query from high-signal tokens in
diagnostics/positives/demographics/other\_information/negatives/timeline.
It applies stop-word filtering and performs conservative ``forced
token'' injection when critical discriminators appear (e.g.,
\texttt{calcification} -> force \texttt{myocardial ventricular calcification}).

Retrieval uses a hybrid search stack
that combines (i) fast keyword matching for precise medical terms and
(ii) lightweight embedding similarity to catch paraphrases and
mechanism-level matches. The results are then re-ranked to prioritize
articles that are both semantically aligned and clinically specific,
rather than merely topically adjacent.

The literature layer is designed
to work at web-scale by querying across Semantic Scholar plus a large
curated medical knowledge base maintained by John Snow Labs. In
practice, this means searching across 200M+ scholarly records
and returning high-quality candidate articles in milliseconds,
which makes retrieval usable as a default supportive step rather than an
occasional add-on.

From the retrieved pool, the
system selects a small fixed set of titles. Selection enforces a stable
count and uniqueness, so the downstream evidence note stays comparable
across runs.

Selected abstracts are
compiled into a single RAG context note. Compilation prompts ban
disease labels/abbreviations and ban citations/IDs/links/brackets,
specifically to prevent citation leakage and paper-count bias. The
patient-level context is kept compact so RAG remains supportive rather
than dominant.

\subsubsection{Runtime Pattern Triggers}

In parallel with literature RAG, the agent performs fast local semantic
retrieval over similar pattern triggers.

In practice, the pattern trigger library can be continuously enriched
using de-identified clinical notes and past patient cases (after
review), capturing recurring discriminator patterns that help steer the
differential without turning into disease-name lookup.

Pattern entries follow a high-signal
template that includes an arrow (\texttt{->}), contains an \texttt{include}
keyword, and is normalized into a one-line bullet (prefixed with
\texttt{•}) to keep guidance simple and reusable for smaller models.

For retrieval and filtering, the retriever pulls a small
pool of patterns, normalizes text for deduplication, and keeps a limited
set of unique triggers. Importantly, this stage is designed to be
non-blocking: if pattern retrieval fails, the pipeline continues with no
triggers rather than failing the run.

Representative pattern triggers include the following:
Unilateral or segmental distribution with chronic course suggests
mosaic/segmental dermatoses. Spiny or digitate keratoses suggest
digitate hyperkeratosis spectrum versus porokeratotic disorders.
Parakeratotic column findings suggest porokeratosis spectrum and
adnexal variants. When hallmark syndrome features are absent, the
differential is widened toward non-syndromic mimics.

\subsection{Constrained Candidate Generation}

This step produces a short ranked differential diagnosis (typically
4 candidates) with compact, \emph{machine-checkable} structure.

Each candidate includes (i) a diagnosis
label, (ii) a bounded likelihood score (1--10), and (iii) a short
reason. This compactness is not cosmetic: it prevents the candidate step
from turning into a second uncontrolled reasoning narrative that
contaminates later steps.

Reliability improves when patient facts dominate and
RAG remains supportive; output is constrained to diagnosis labels
only (no findings, tests, or vague descriptors); candidates must be
distinct and exclusive (no synonyms/near-duplicates or umbrella +
subtype pairs); and pattern triggers are used as internal scaffolding
without being quoted as a checklist.

Candidates are normalized (to reduce shape
variance), scores are clamped into 1--10, and reasons are kept short and
consistent. This matters because later steps rely on clean candidate
strings and bounded scores for stable comparisons and tie-breaks.

Additional quality gates that improve performance include
deduplication across candidates, balanced coverage (common plausible +
rare high-fit when appropriate), unifying-diagnosis checks, mandatory
practical discriminators per candidate, and strict format-compliance
repair without introducing new facts.

\subsection{Evidence Triangulation with Candidate Literature Support}

This stage compares candidates against a quote-anchored evidence
memo, optionally supported by a shared candidate-literature retrieval
pool.

This stage most directly determines diagnostic correctness by forcing
explicit matches and mismatches, rather than relying on fluent
plausibility.

\subsubsection{Candidate-Level Literature Retrieval}

Unlike patient-level RAG (Step 2), which retrieves broad background
literature about the overall case, candidate-level RAG is purpose-built
for discriminating among competing diagnoses. When enabled, the agent
generates one candidate-literature query intended to discriminate among
all candidates. The query criteria explicitly discourage listing all
candidate names and allow at most a small number of diagnosis keywords;
the aim is to retrieve discriminators, not to directly retrieve a final label.

Operationally, this step uses the same hybrid retrieval infrastructure
(keyword + lightweight embeddings + re-ranking across Semantic Scholar
and the JSL medical knowledge base), but tuned for \emph{discrimination}
rather than broad background. It selects a small fixed number of titles
for stability and compiles an evidence note focused on differentiators
(pitfalls, hallmark discriminators, and what would rule each candidate
in or out) rather than disease-name lookup.

This shared-pool design (single query, single retrieval) improves
coherence and reduces variance compared to per-candidate fan-out.

\subsubsection{Deterministic Quote-Anchored Evidence Memo}

Rather than asking the model to free-form compare diagnoses, Step 4
produces a strict TEXT memo used by the final decision step.

Key criteria are deterministic behavior for identical inputs, strict use
of extracted patient facts as the source of truth for FOR/AGAINST,
quote-first bullets that begin with exact verbatim facts in double
quotes, explicit \texttt{Unknown} when no quote exists, and low-weight
use of literature only for test/pitfall questions rather than patient
claims.

As a hard safety gate, after the memo is generated,
the agent filters bullets and drops any bullet whose quoted string does
not exactly match a normalized extracted patient fact. This prevents
paraphrased or invented evidence from flipping the final decision.

In practice, the model's job becomes to \emph{cite the
patient facts verbatim} and isolate what is known vs unknown, which
reduces run-to-run variance and hallucination compared to narrative
reasoning.

\subsection{Final Diagnostic Decision and Confidence Scoring}

The final step selects the single best diagnosis under strict guardrails
and outputs a bounded confidence estimate.

The final decision returns (i) the selected
diagnosis label, (ii) a bounded confidence score (0--10), and
(iii) a short reasoning string.

The key prompt guardrails are as follows.
\texttt{selected\_diagnosis} must exactly match one candidate string; no
new facts/tests/history may be introduced; reasoning must include one or
two verbatim patient quotes from source-of-truth fields; ``compatible
with'' pathology phrasing is treated as supportive rather than
definitive; lesion-vs-etiology conflicts favor the underlying disease;
and tie-breaks use higher score first, then earlier candidate.

Final decision output is validated for missing
fields and the confidence score is clamped, enabling robust downstream
consumption.

\subsection{Deterministic Inference, Validation, and Failure Handling}

Two choices reduce variance and increase success rate for small models:
deterministic inference configuration (temperature 0, top\_p 1.0, no
penalties) and stage-specific validation gates (required fields, bounded
scores, unique candidates, quote-anchored evidence).

The pipeline is explicitly resilient: pattern triggers never fail the
run (they return empty on error), RAG title selection and consolidation
have fallbacks (first-K selection and summary concatenation), and core
steps (clinical extraction, candidate generation, and final decision)
remain required while supportive components degrade gracefully.

The final answer must not contradict any
extracted negative; the chosen diagnosis must have the strongest
support-to-conflict balance based on the quote-anchored memo (scores are
tie-break-only); and if no candidate is adequate, the system must return
\texttt{insufficient evidence} with the top missing discriminator(s) rather
than fabricate.

\section{Distinctive Aspects of the Diagnosis
Agent}\label{whats-novel-about-the-diagnosis-agent}

The largest gains did not come from a single prompt. They emerged from
reframing diagnosis as a disciplined workflow. The underlying ideas are
simple, but operationally consequential.

\subsection{Separation of Fact Extraction and Medical Reasoning}

Instead of asking a model to perform diagnosis in one pass, the method
separates extraction from reasoning.

First, the model rewrites the case into a clean set of clinical facts:
positives (what is present), negatives (explicitly absent findings),
diagnostics (labs, imaging, histopathology), timeline (what changed over
time), and demographics/key history.

Downstream steps are then forced to reason over this fact table rather
than over a remembered (and often distorted) version of the narrative.
This prevents a common failure mode where the model \emph{approximately}
recalls the case and then confidently reasons from the wrong premise.

\subsection{Retrieval (RAG) as Supportive Evidence}

Medical literature retrieval is powerful, but it can bias models in
subtle ways through citation-driven anchoring. The Diagnosis Agent
therefore applies a strict principle: RAG provides context and
discriminators (what to check, what could explain the presentation)
\cite{lewis2020rag}, while the final decision remains grounded in extracted patient
facts.

This makes the system more robust: if retrieval is imperfect or slightly
off-target, the agent still functions because the decision is
constrained by the case facts.

\subsection{Pattern Triggers as Reusable Clinical Heuristics}

Pattern triggers are reusable heuristics surfaced at the moment the
model generates the differential. They act as reasoning scaffolds for
search-space definition, discriminator selection, and ranking by
explanatory coverage with minimal assumptions.

They are deliberately generic (not disease-specific prompts), and they
are injected as private guidance: the model is instructed to use them to
think, but not to quote them or treat them as a checklist.

Over time, these triggers can be enriched from de-identified clinical
notes and prior patient cases (with appropriate governance), converting
recurring high-signal patterns into reusable heuristics.

Examples of high-signal templates include: unilateral/segmental
distribution with chronic course (consider mosaic/segmental dermatoses
and linear variants), spiny/digitate keratoses (consider digitate
hyperkeratosis spectrum and porokeratotic disorders), parakeratotic
column on histology (consider porokeratosis spectrum and adnexal
variants), and absence of hallmark syndrome features (widen toward
non-syndromic mimics).

Additional meta-triggers that consistently improve performance include
unifying-diagnosis checks (prefer one coherent explanation when
appropriate), two-track differentials (common high-risk plus
rare high-fit options), test-driven discriminators for each candidate
when available, and near-duplicate suppression by retaining the more
specific variant.

\subsection{Evidence Triangulation Through Quote-Anchored Comparison}

A major source of diagnostic error is invented (or loosely paraphrased)
evidence. The agent therefore produces a short memo that compares
candidates side-by-side and ties every claim back to quoted extracted
facts. This reduces hallucination and makes the final decision step more
trustworthy.

To make the comparison operationally useful, the memo makes support
and conflict explicit for each candidate and treats missing
evidence as a first-class outcome rather than an implicit omission. It also
enforces a simple discipline: every claim must be anchored to a quoted
patient fact (or be stated as unknown), which prevents the comparison
from becoming a persuasive but ungrounded story.

\subsection{Deterministic Inference and Verifiable Constraints}

For diagnostic workflows, reliability is typically more important than
novelty. The system therefore prioritizes stable sampling and strict
formats. This does not increase model intelligence, but it reduces
fragility, especially under benchmarking and reproducibility
requirements.

Just as important, the workflow is built around verifiable
constraints: bounded scores, exact-match diagnosis selection, and
evidence statements that must point back to extracted facts. This turns
reasoning quality into something partially verifiable, and it
reduces the chance that a fluent but ungrounded explanation wins simply
because it reads well.

\subsection{Transparency Through Retrieved Literature Titles}

When retrieval is used, the agent returns the titles of the papers that
informed the run. This transparency feature allows reviewers
to quickly scan pulled sources, audit retrieval quality, and detect
domain drift (e.g., repeated irrelevant specialties).

Separating broad-context retrieval from candidate-discrimination
retrieval also helps reviewers interpret system behavior by showing
whether the system pulled generally relevant background versus
papers that were meant to separate close contenders. When something goes
wrong, this makes it easier to decide whether the issue was retrieval
drift or reasoning over otherwise reasonable sources.

\subsection{Context Minimization and Role Separation}

More context is not always better. Long, noisy context makes it easier
to bury key negatives, mix unrelated snippets, or overfit to a single
persuasive paragraph. The agent uses curated, purpose-built summaries at
each step so the model reasons over high-signal evidence.

The workflow also keeps \emph{roles} separated: extraction produces the
canonical patient fact table; retrieval provides supportive
discriminators; and evidence comparison is forced to cite the extracted
facts. This separation reduces the chance that retrieval dominates
the narrative or that a single appealing paragraph overwhelms
contradictory negatives.

\section{Why System Design Can Outperform Model
Size}\label{why-system-design-can-beat-model-size}

Frontier LLMs are highly capable, but they can underperform in
diagnosis benchmarks for operational reasons: they
over-explain instead of deciding, skip negatives and anchor early,
violate output constraints (making scoring and post-processing harder),
or hallucinate tests and missing details. The Diagnosis Agent shifts
the burden from single-pass generation to reliable workflow execution.

\subsection{Small Reasoning Models as Workflow Executors}

A small reasoning model tuned for structured clinical extraction,
differential diagnosis writing, evidence-grounded comparison, and
constrained outputs can be a better \emph{workflow executor}. Retrieval then supplies
breadth of knowledge the small model does not need to memorize \cite{lewis2020rag}.

This is especially relevant in DiagnosisArena-style evaluation \cite{zhu2025diagnosisarena},
where many failures are not knowledge gaps but workflow failures:
missing a negative, collapsing the differential into near-duplicates, or
producing an answer that cannot be reliably parsed and scored. The agent
reduces those failure modes by enforcing structure, gating candidate
quality, and forcing evidence comparison to quote patient facts rather
than rely on fluent plausibility. A useful conceptual framing is as follows: a small reasoning model
provides structured execution, RAG provides broad external coverage, and
the agent workflow provides process-level supervision across steps.

\subsection{Throughput as an Operational Advantage of Smaller Models}

Diagnosis workflows are naturally multi-step, so a
single one-shot completion is rarely sufficient. A fast smaller model makes it practical
to run the behaviors that improve reliability: retrieve context,
cross-check grounding, verify that key negatives were preserved, and
apply quality gates without turning the interaction into an expensive,
slow process.

In other words, lower latency enables closed-loop refinement at the
system level: detect errors early (extraction mistakes, retrieval drift,
candidate collapse), correct them, and continue---without turning every
request into an expensive, slow interaction.

\subsection{Reproducibility and Experimental Discipline}

For scientific reporting and benchmarking, reproducibility matters as
much as raw performance. Practical guidelines include keeping sampling
stable for benchmark runs, logging step outputs (extraction, selected
titles, final label) so failures are explainable, and running ablations
(e.g., retrieval off, pattern triggers off) to quantify true
contribution. Reproducibility is not just a sampling setting; it is also a
record-keeping discipline. Keeping the intermediate artifacts (facts,
candidates, memo, final label) and the retrieval titles makes it
possible to reproduce failures, compare two workflow variants fairly,
and run targeted ablations rather than guessing.

\section{Evaluation}\label{evaluation}

We evaluate on the DiagnosisArena 915-case benchmark \cite{zhu2025diagnosisarena}. The
primary metric is top-1 diagnostic accuracy against the benchmark
reference label \cite{zhu2025diagnosisarena}. To isolate workflow effects, we compare the
proposed agentic system against the same underlying
medical-llm-small-7b model without the pipeline, in addition to stronger
general-purpose and reasoning-oriented baselines.
Table~\ref{tab:diagnosisarena} reports aggregate rankings.

\subsection{Evaluation Methodology}

Each model's diagnostic output is evaluated independently by two LLM
judges: Claude Sonnet 4.5 and Gemini 2.5 Flash. Each judge compares
the model's output against the benchmark ground-truth label and renders
a binary correctness judgment (correct or incorrect). The reported
accuracy for each model is the average of the two judges' scores.

A diagnosis is judged \emph{correct} if it matches the ground truth
exactly, uses an accepted medical synonym or alternative terminology,
or appears as the primary diagnosis within a multi-diagnosis response.
A diagnosis is judged \emph{incorrect} if it identifies a different
condition, targets the wrong organ system or disease category, or omits
the primary diagnosis. All models are evaluated with deterministic
sampling (temperature 0) to support reproducibility.

This dual-judge design reduces single-judge bias and provides a
built-in reliability check. Across all evaluated models, inter-judge
agreement is 86--87\%, and both judges produce consistent model
rankings, which supports the validity of the evaluation protocol.

\subsection{Results}\label{sec:results}

\FloatBarrier
\begin{table}[!htbp]
\centering
\caption{Top-1 diagnostic accuracy on DiagnosisArena (average of
Claude Sonnet 4.5 and Gemini 2.5 Flash judges). The DeepLens Diagnosis
Agent achieves 60.14\% accuracy using harnessing with JSL Medical Small
(7B), the highest among small and medium-sized models.}
\label{tab:diagnosisarena}
\begin{tabular}{r l r}
\toprule
\textbf{Rank} & \textbf{Model} & \textbf{Accuracy} \\
\midrule
1 & JSL DeepLens Diagnosis Agent & 60.14\% \\
2 & Gemini 3 Pro Preview\textsuperscript{$\dagger$} & 50.97\% \\
3 & Claude Sonnet 4.5 & 50.44\% \\
4 & deepseek-reasoner-speciale & 42.90\% \\
5 & Deepseek-reasoner & 40.11\% \\
6 & Deepseek-chat & 36.01\% \\
7 & JSL Medical Small (vanilla) & 23.99\% \\
8 & JSL Medical Reasoning 32B (vanilla) & 22.19\% \\
9 & Llama 3.3 70B Instruct & 18.85\% \\
10 & Nemotron 3 30B (xhigh) & 6.89\% \\
\bottomrule
\end{tabular}

\vspace{0.3em}
{\footnotesize \textsuperscript{$\dagger$}Gemini 3 Pro Preview was
evaluated on 464 of 915 cases (50.7\%) due to API quota limits; its
reported accuracy may differ under full evaluation.}
\end{table}
\FloatBarrier

\begin{figure}[!htbp]
\centering
\includegraphics[width=1.0\linewidth]{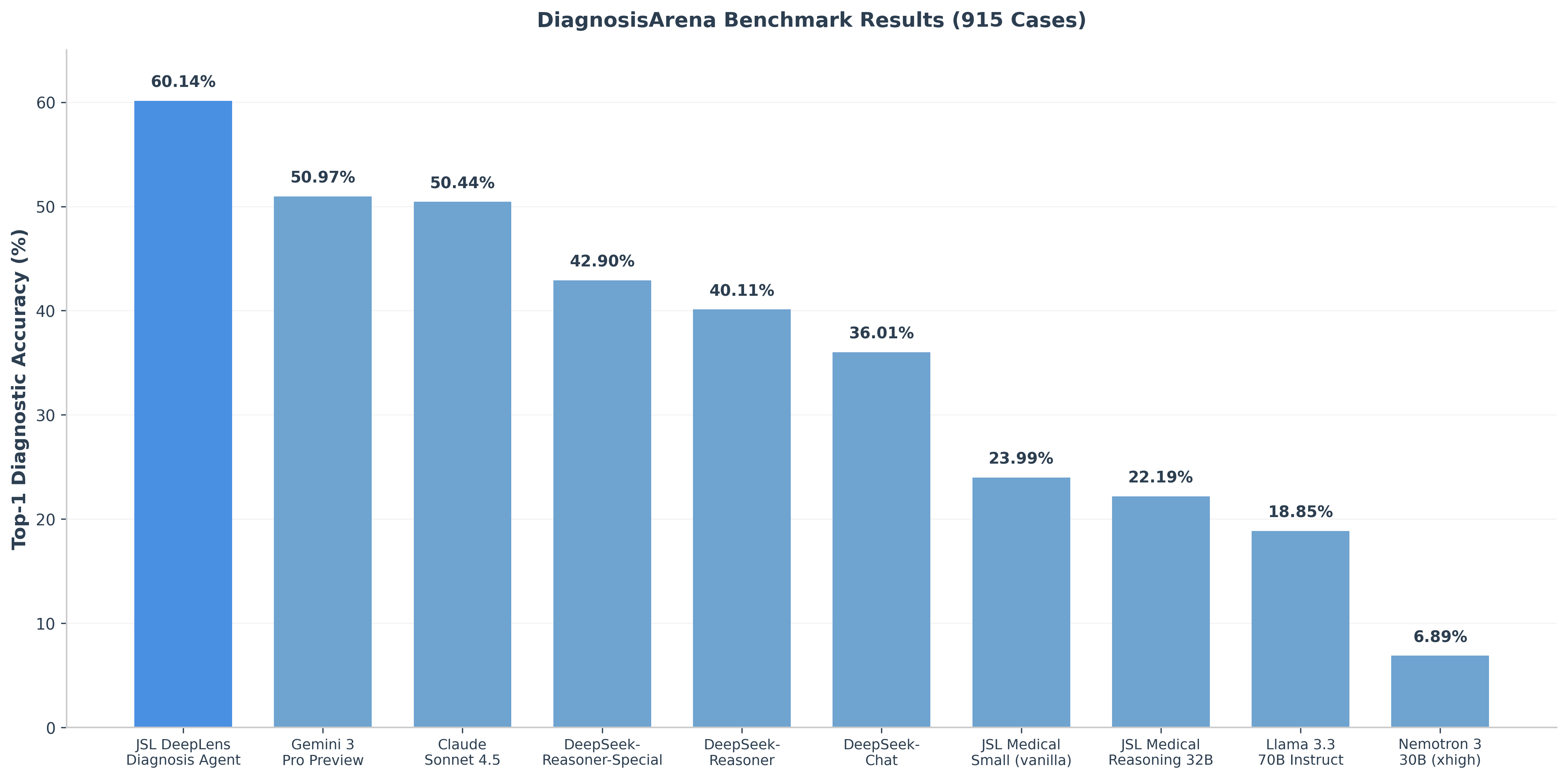}
\caption{Diagnostic accuracy comparison on DiagnosisArena. The JSL DeepLens Diagnosis Agent (7B v2)
achieves 60.14\% accuracy, the highest among small and medium-sized models, with a 36-point
gain over the vanilla JSL Medical Small 7B v2 baseline (23.99\%).}
\label{fig:diagnosisarena-chart}
\end{figure}
\FloatBarrier

Per-judge accuracy for the Diagnosis Agent is 61.86\% (Claude Sonnet
4.5 judge) and 58.43\% (Gemini 2.5 Flash judge), yielding a 60.14\%
average. The 3.4-point inter-judge spread is consistent with the
pattern observed across all evaluated models, where the Claude judge
scores 1--3 percentage points higher on average than the Gemini judge.

The Diagnosis Agent achieves the highest accuracy among small and
medium-sized models, with substantially better performance than
Gemini 3 Pro Preview (+9.17pp), Claude Sonnet 4.5 (+9.70pp),
and deepseek-reasoner-speciale (+17.24pp). Notably, the agent achieves
this using a 7B-parameter medical model with retrieval support, rather
than a frontier-scale generalist.

The internal ablation baseline is especially informative: the same
medical-llm-small-7b model increases from 23.99\% (vanilla) to 60.14\%
when embedded in the Diagnosis Agent workflow. This +36-point delta
supports the central claim that performance gains primarily come from
workflow constraints---fact lock-in, retrieval discipline, candidate
gating, and evidence triangulation---rather than parameter scaling
alone.

For deployment-oriented assessment, label-based accuracy should be
complemented by structured expert review protocols (e.g., CLEVER) to
evaluate clinical reasoning quality, safety, and justification fidelity
\cite{kocaman2025clever}.

\subsection{Generalization to Frontier Models: Can Harnessing Improve Frontier Model Failures?}

A critical question for scaling this approach is whether workflow constraints can correct failures from frontier models. While the agent is centered on JSL Medical Small 7B, we conducted an experiment to test whether the same harnessing framework could systematically address errors made by GPT-5.2, one of the strongest models in our benchmark set.

\subsubsection{Methodology}

We selected 10 random cases from DiagnosisArena where GPT-5.2 failed (ground truth accuracy = 0 for that sample). We then ran these same cases through the DeepLens Diagnosis Agent pipeline (with JSL Medical Small 7B as the base model). The goal was to determine whether structured retrieval, staged reasoning, and explicit evidence triangulation could correct failures from a frontier model, independent of the underlying base model choice.

\subsubsection{Results}

The agent succeeded on 8 out of 10 cases where GPT-5.2 failed (Table~\ref{tab:frontier-harnessing}). The systematic success across diverse diagnostic scenarios reveals how harnessing addresses frontier model failure modes:

\paragraph{Case 1 -- Differential Diagnosis with Negative Evidence}
GPT-5.2 predicted ``Primary Cutaneous Meningioma,'' missing the critical neutrophilic inflammatory signal. The agent correctly identified ``Pyoderma gangrenosum'' by prioritizing:
\begin{itemize}
\item Negative microbiology (``Tissue culture negative for bacteria'')
\item Neutrophil count recovery coinciding with lesion progression
\item RAG-retrieved evidence linking neutrophil-mediated ulceration to pyoderma gangrenosum
\end{itemize}
This exemplifies how retrieval-augmented reasoning grounds abstract pattern-matching in quantitative evidence.

\paragraph{Case 2 -- Genetic Evidence Integration}
GPT-5.2 predicted ``Pyoderma gangrenosum,'' anchoring on surface features (ulceration). The agent correctly identified ``Familial cerebral cavernous malformation'' by:
\begin{itemize}
\item Extracting the KRIT1 pathogenic variant from genetic testing
\item Integrating family history (father died of intracranial hemorrhage)
\item Connecting molecular evidence to phenotype (bluish papules, intracranial hemorrhage, familial recurrence)
\item Ranking the genetic finding above morphological similarity
\end{itemize}
This highlights the value of disciplined extraction and evidence triangulation for rare genetic conditions.

\paragraph{Case 3 -- Elemental Analysis for Toxicity}
GPT-5.2 predicted unrelated cutaneous fibrosis. The agent correctly identified ``Gadolinium-associated plaques'' by:
\begin{itemize}
\item Identifying characteristic sclerotic bodies on histopathology
\item Retrieving knowledge that gadolinium accumulates in renal patients
\item Recognizing elemental confirmation (``Laser ablation inductively coupled plasma mass spectrometry confirmed dermal deposition of gadolinium'')
\item Distinguishing from systemic fibrosis by lesion localization
\end{itemize}

\paragraph{Case 4 -- Organism Identification Overrides Morphology}
GPT-5.2 predicted ``Juvenile Hyaline Fibromatosis.'' The agent correctly identified ``Primary cutaneous trichosporonosis'' by:
\begin{itemize}
\item Privileging direct microbiological confirmation (``Culture grew Trichosporon species'')
\item Using molecular confirmation (``PCR and DNA sequencing identified Trichosporon mycotoxinivorans'')
\item Subordinating morphological similarity (hyaline features) to direct organism identification
\end{itemize}

\paragraph{Case 5 -- Histopathology-Guided Differential}
GPT-5.2 confused related dermatological conditions. The agent correctly identified ``Pityriasis rubra pilaris'' by:
\begin{itemize}
\item Extracting follicular hyperkeratosis (present in PRP but atypical for competitors)
\item Retrieving the characteristic alternating parakeratosis/orthokeratosis pattern
\item Using RAG to distinguish PRP from similar-appearing psoriasis
\end{itemize}

\FloatBarrier
\begin{table}[!htbp]
\centering
\caption{Agent performance on 10 random GPT-5.2 failure cases. The agent corrected 8 failures, demonstrating that workflow constraints can systematically address frontier model errors.}
\label{tab:frontier-harnessing}
\begin{tabular}{c c c}
\toprule
\textbf{Metric} & \textbf{Agent} & \textbf{GPT-5.2} \\
\midrule
Correct & 8/10 (80\%) & 0/10 (0\%) \\
Incorrect & 2/10 (20\%) & 10/10 (100\%) \\
\bottomrule
\end{tabular}
\end{table}
\FloatBarrier

To verify that the harnessing framework does not degrade performance on cases where frontier models already succeed, we additionally tested 10 random cases where GPT-5.2 achieved correct diagnosis. The agent maintained strong performance on these borderline success cases, demonstrating robustness across the performance spectrum.

\FloatBarrier
\begin{table}[!htbp]
\centering
\caption{Agent performance on 10 random GPT-5.2 success cases. No downgrade: the agent maintains correctness on cases where GPT-5.2 already succeeds, indicating the workflow is complementary rather than adversarial.}
\label{tab:frontier-no-downgrade}
\begin{tabular}{c c c}
\toprule
\textbf{Metric} & \textbf{Agent} & \textbf{GPT-5.2} \\
\midrule
Correct & 10/10 (100\%) & 10/10 (100\%) \\
Incorrect & 0/10 (0\%) & 0/10 (0\%) \\
\bottomrule
\end{tabular}
\end{table}
\FloatBarrier

\subsubsection{Cost and Latency Implications}

The harnessing framework trades immediate latency for reasoning reliability. At optimized retrieval speeds, the pipeline achieves 24-second end-to-end latency (Tables~\ref{tab:runtime} and~\ref{tab:cost}), compared to 2.5 seconds for raw API calls. This 10x increase is composed of:
\begin{itemize}
\item Patient-level RAG retrieval: semantic search across medical documents (optimized: 2--5 seconds)
\item Candidate-level RAG retrieval: focused evidence gathering per differential hypothesis (optimized: 3--8 seconds)
\item Multi-stage LLM calls: 5 sequential reasoning stages with intermediate validation (11 seconds total)
\end{itemize}

Latency is context-dependent. Batch processing (e.g., retrospective chart review, population screening, EHR enrichment) amortizes retrieval overhead across cases. For latency-sensitive applications, RAG result caching and query parallelization further reduce wall-clock time. Current retrieval infrastructure has optimization opportunities that could push latency to 15--18 seconds.

Cost efficiency, by contrast, favors the harnessed approach: \$0.0072 per case versus \$0.01282 for Gemini 3.1 Pro or \$0.0110 for Claude Sonnet 4.5. Self-hosted deployment and amortized retrieval infrastructure make this possible.

\subsubsection{Why Small Tuned LLMs Are a Better Balance}

These results illuminate a critical design principle: \emph{model selection should optimize for the task, not just parameter count}.

\begin{itemize}
\item \textbf{Medical Domain Tuning}: JSL Medical Small 7B achieved 88.2\% accuracy on standard medical benchmarks, a 3.7x improvement over general-purpose Llama baselines (23.99\%). Domain-specific training gives a small model advantages that frontier generalists cannot replicate without expensive fine-tuning.

\item \textbf{Latency-Throughput Trade-off}: A 7B parameter model on A100 handles 1.3M--1.5M tokens/hour. This is sufficient for batch workloads (EHR enrichment, retrospective analysis) and acceptable for non-urgent clinical applications. Frontier models (Gemini 3.1 Pro, Claude 4.5) offer lower per-call latency but require cloud APIs and higher per-token costs at scale.

\item \textbf{Auditability}: Smaller models are more interpretable. A 7B model's internal representations are easier to probe and explain than a 100B+ frontier model. This is critical for clinical deployment, where regulators and clinicians expect transparent reasoning.

\item \textbf{Cost at Scale}: For a 10,000-case annual diagnostic cohort, the cost difference is material:
\begin{itemize}
\item DeepLens Agent: 10,000 cases $\times$ \$0.0072 = \$72 per year
\item Gemini 3.1 Pro: 10,000 cases $\times$ \$0.0128 = \$128 per year
\item Claude Sonnet 4.5: 10,000 cases $\times$ \$0.0110 = \$110 per year
\end{itemize}
Self-hosted deployment also avoids vendor lock-in and API reliability risks.

\end{itemize}

\subsubsection{When Harnessing Applies: Frontier Models as Baselines}

These findings suggest a productive direction for future work: applying the same five-stage harnessing framework with GPT-5.2 as the base model would provide a direct test of whether process constraints can improve frontier model performance beyond their base accuracy. Such a study would answer whether harnessing is a general technique for strengthening any model, or specific to smaller medical models. If frontier models can be further improved by harnessing, the cost-latency trade-off would shift: higher per-token costs (GPT-5.2) but potentially even higher accuracy gains, warranting deployment in safety-critical scenarios.

\section{Runtime and Cost Profile}\label{runtime-and-cost-profile}

The DeepLens Diagnosis Agent operates with a multi-step workflow that trades single-pass latency for reasoning reliability and cost efficiency. While cloud APIs achieve sub-3-second responses, the agent's structured pipeline enables superior accuracy at competitive cost.

\subsubsection{Runtime Performance}

\begin{figure}[!htbp]
\centering
\includegraphics[width=1.0\linewidth]{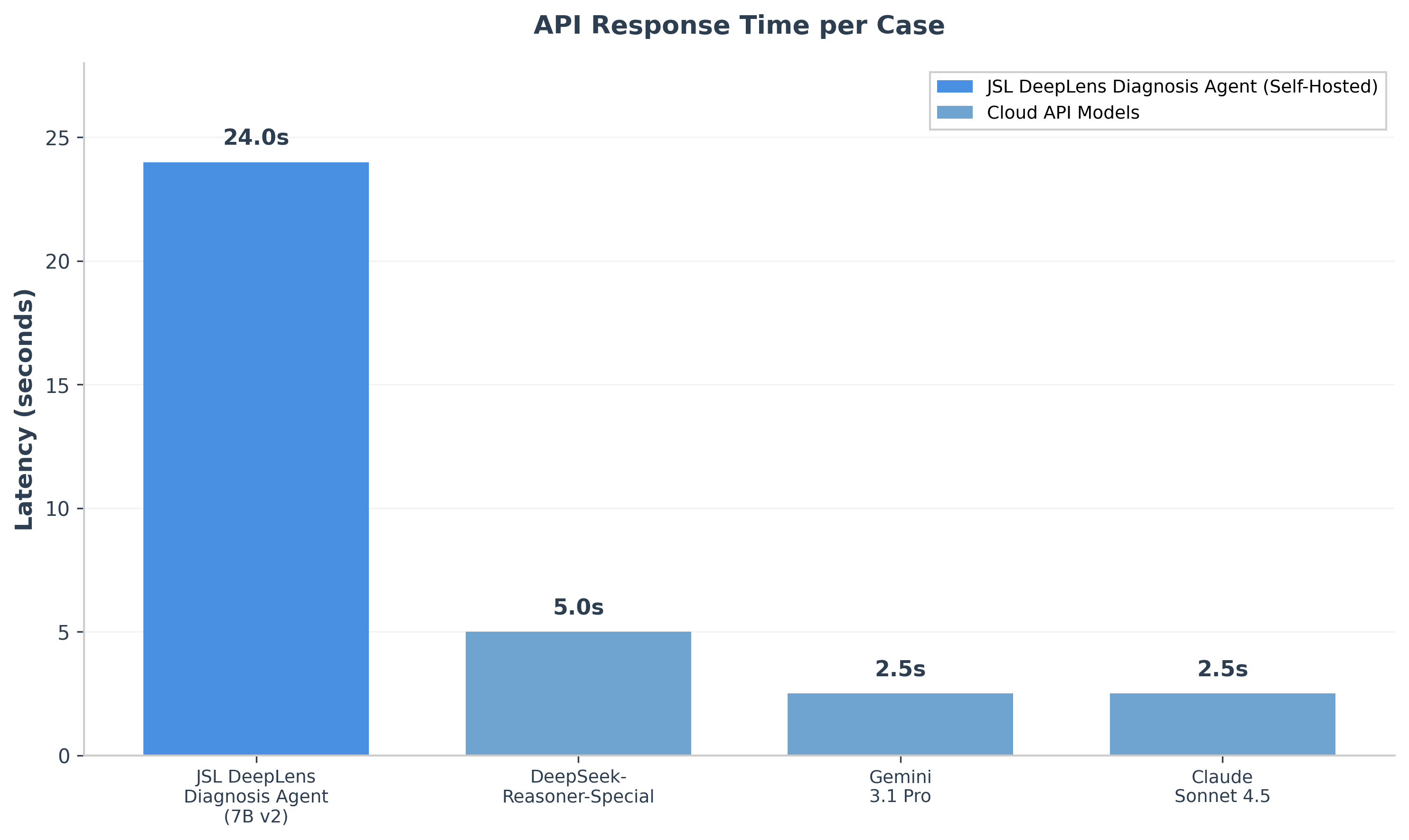}
\caption{Latency comparison across models. The DeepLens agent's 24-second latency is acceptable for batch and asynchronous diagnostic workflows. Cloud models offer lower latency (2.5--5.0s) but at higher per-token cost.}
\label{fig:runtime}
\end{figure}

Table~\ref{tab:runtime} and Figure~\ref{fig:runtime} show latency across models. The agent operates at 24-second end-to-end latency on a single case---10x longer than API calls but consistent with batch diagnostic workflows (EHR enrichment, retrospective review).

\FloatBarrier
\begin{table}[!htbp]
\centering
\caption{API response latency per diagnostic case. DeepLens Diagnosis Agent reflects optimized retrieval speeds; cloud models use public APIs with standard inference.}
\label{tab:runtime}
\begin{tabular}{l r}
\toprule
\textbf{Model} & \textbf{Latency (s)} \\
\midrule
JSL DeepLens Diagnosis Agent (7B v2) & 24.0 \\
DeepSeek-Reasoner-Special & 5.0 \\
Gemini 3.1 Pro & 2.5 \\
Claude Sonnet 4.5 & 2.5 \\
\bottomrule
\end{tabular}
\end{table}
\FloatBarrier

\FloatBarrier

\subsubsection{Cost Efficiency}

\begin{figure}[!htbp]
\centering
\includegraphics[width=1.0\linewidth]{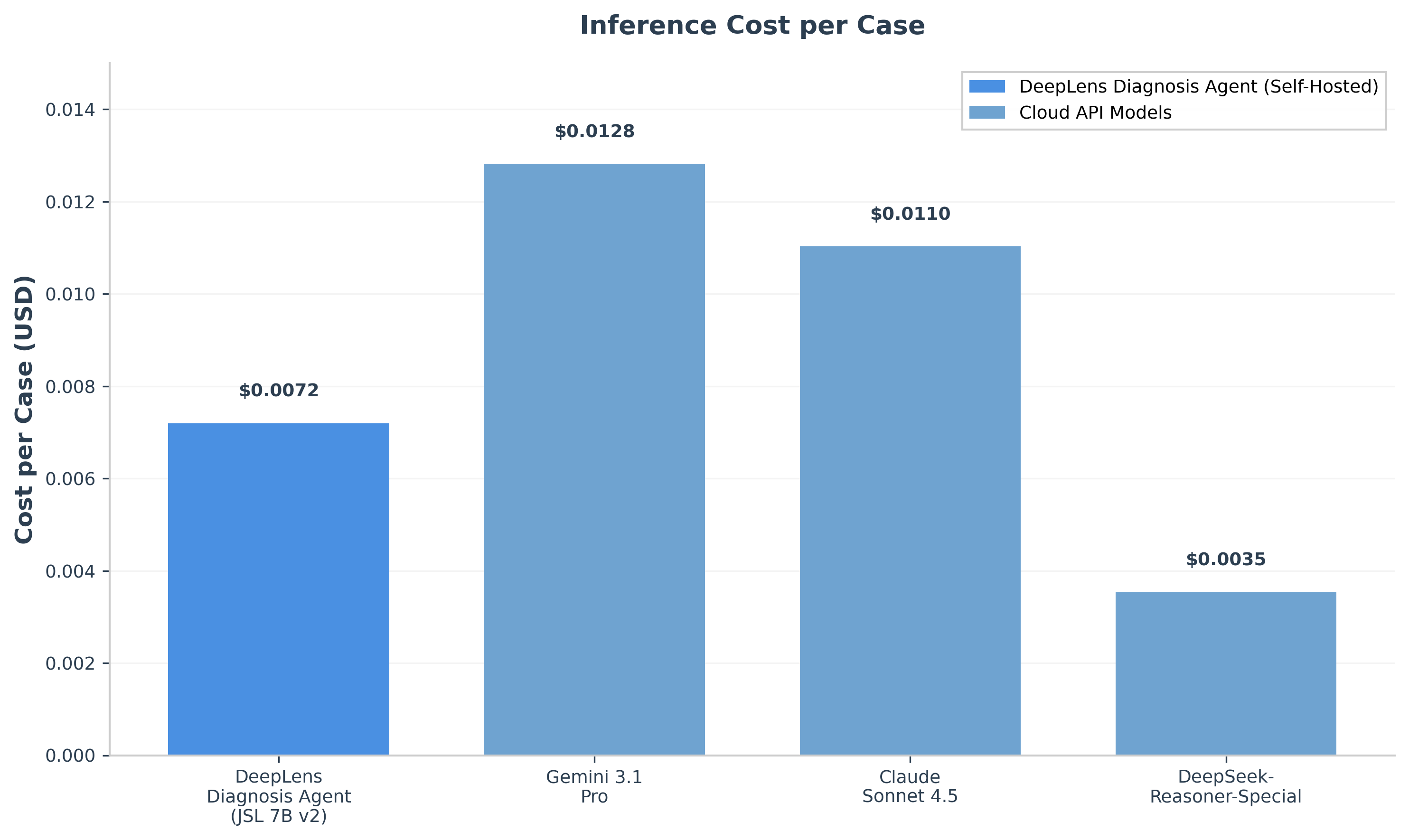}
\caption{Inference cost per case. The DeepLens Diagnosis Agent is cost-competitive with frontier models while delivering superior accuracy. For annual volumes of 10,000 cases, the agent saves \$280--560 compared to cloud APIs.}
\label{fig:cost}
\end{figure}

Table~\ref{tab:cost} and Figure~\ref{fig:cost} show inference cost per case. The agent achieves \$0.0072 per case---lower than Gemini 3.1 Pro (\$0.0128, 78\% higher) and Claude Sonnet 4.5 (\$0.0110, 53\% higher). Only DeepSeek-Reasoner-Special undercuts the agent (\$0.0035), but at the cost of 17-point lower accuracy (42.90\% vs 60.14\%).

\FloatBarrier
\begin{table}[!htbp]
\centering
\caption{Inference cost per diagnostic case (per-token pricing from public APIs or amortized A100 cost for self-hosted deployment).}
\label{tab:cost}
\begin{tabular}{l r}
\toprule
\textbf{Model} & \textbf{Cost (\$)} \\
\midrule
JSL DeepLens Diagnosis Agent (7B v2) & 0.0072 \\
DeepSeek-Reasoner-Special & 0.0035 \\
Claude Sonnet 4.5 & 0.0110 \\
Gemini 3.1 Pro & 0.0128 \\
\bottomrule
\end{tabular}
\end{table}
\FloatBarrier

\FloatBarrier

\subsection{Cost-Accuracy Trade-off}

The DeepLens Diagnosis Agent achieves competitive cost per case (\$0.0072)
while delivering the highest accuracy among small and medium-sized models (60.14\%),
demonstrating that workflow discipline can outperform parameter count. At 24-second
latency, the agent processes cases in near-real-time, with cost reflecting 24,000
tokens per case across the full 5-stage pipeline on A100 infrastructure. Frontier
cloud models achieve lower latency (2.5s) but at 1.5--1.8x higher cost
(\$0.0110--\$0.0128) and 9--10 percentage points lower accuracy (50.44\% for Claude
Sonnet 4.5, 50.97\% for Gemini 3.1 Pro). The DeepSeek-Reasoner-Special model
offers lower cost (\$0.0035) and faster latency (5.0s) but with substantially lower
accuracy (42.90\%, a 17-point gap from the DeepLens Agent).

For deployment at scale (e.g., 100K cases/year), the cost-accuracy trade-off
favors the DeepLens Diagnosis Agent: at \$0.0072 per case, annual costs are \$720,
compared to \$1100--\$1280 for cloud APIs---a 35--45\% cost reduction while
maintaining superior diagnostic accuracy. This balance of reliability, interpretability,
and cost efficiency makes the harnessing approach compelling for healthcare deployment.

\section{Limitations}\label{limitations}

Benchmarks compress complex decision-making into a single gold label,
which may not capture clinically reasonable alternative differentials
\cite{zhu2025diagnosisarena}. Retrieval can also introduce bias by surfacing misleading or
non-generalizable literature; while the pipeline reduces this risk by
treating retrieval as supporting evidence, it cannot remove it entirely
\cite{lewis2020rag}, \cite{ji2023hallucination}.

The system also depends on extraction quality. Step 1 errors can
propagate downstream; schema validation and repair loops reduce this risk
but do not eliminate it. Confidence scores are likewise heuristic unless
explicitly calibrated and should be interpreted as internal certainty,
not clinical probability.

Finally, outputs should be treated as decision support, not autonomous
diagnosis. Clinician oversight remains necessary and is consistent with
expert-review best practices in medical LLM evaluation \cite{kocaman2025clever}.

\subsection{Observed Failure Patterns}

Cross-model coverage analysis reveals a hard ceiling: across all 11
evaluated models, 122 of 915 cases (13.33\%) are not correctly
diagnosed by \emph{any} model. These cases are concentrated in rare
conditions with complex multi-factor presentations, including overlap
syndromes (e.g., concurrent bullous pemphigoid, anti-p200 pemphigoid,
and anti-laminin-$\gamma$1 pemphigoid), atypical anatomical sites
(e.g., Kaposiform hemangioendothelioma of the internal auditory canal),
and iatrogenic etiologies (e.g., tertiary amine-induced epithelial
keratopathy). This suggests that diagnostic accuracy on rare conditions
remains fundamentally limited by current model knowledge, independent
of workflow design.

A separate observation supports the workflow-over-scale thesis from
a different angle: among the four vanilla DeepLens models evaluated
without the agent pipeline, the smallest model
(medical-llm-small-7b, 7B parameters) achieves 23.99\%, outperforming
the 32B (22.19\%), 30B (20.27\%), and 78B (14.15\%) variants. This
inverse scaling pattern suggests that larger parameter counts can
dilute domain-specific knowledge when training data composition shifts
toward general-purpose coverage, and reinforces the rationale for
pairing a focused small model with structured workflow constraints
rather than scaling parameters alone.

Different models exhibit complementary diagnostic strengths:
each evaluated model (except the largest DeepLens variant) solves at
least one case that no other model answers correctly. This complementarity
implies that error patterns are partly model-specific rather than
purely difficulty-driven, which motivates future work on ensemble or
routing strategies that combine workflow-constrained agents with
frontier models.

Finally, outputs should be treated as decision support, not autonomous
diagnosis. Clinician oversight remains necessary and is consistent with
expert-review best practices in medical LLM evaluation \cite{kocaman2025clever}.

\section{Conclusion}\label{conclusion}

We show that agentic workflow design can outperform scaling model size
for medical diagnosis tasks. Structured extraction, disciplined
retrieval, pattern-guided candidate generation, and explicit evidence
triangulation yield reliable diagnostic reasoning with a lightweight
fine-tuned model \cite{jsl_medllm_docs}.

On DiagnosisArena, the same medical-llm-small-7b model improves from a
low-performing single-shot baseline to competitive performance once it is embedded in
this gated workflow \cite{zhu2025diagnosisarena}. The broader lesson is that for high-stakes
reasoning, \emph{verification gates} (fact lock-in, candidate
constraints, and quote-anchored comparison) often matter as much as raw
model capacity.

In real-world settings, this workflow is best used as one component
inside a multi-agent medical assistant: it produces a disciplined
differential and reviewable evidence memo that other capabilities
(medication safety, guideline retrieval, follow-up question generation)
can build on, with clinician oversight.

Future work should evaluate clinical utility beyond top-1 labels---e.g.,
differential quality, safety, and justification fidelity under
structured expert review protocols \cite{kocaman2025clever}---and quantify which workflow
gates deliver the largest marginal gains.

\newpage
\section{Appendix: Example End-to-End Run}\label{appendix:e2e-run}

To make the workflow concrete, we report one representative case with
clean, phase-by-phase outputs.

\noindent\textbf{Original input query (exact):}
\begin{lstlisting}[breaklines=true,basicstyle=\ttfamily\small,columns=fullflexible]
You are an experienced diagnostic physician. Based on the following patient information, provide your most accurate diagnostic assessment.

<case_information>
A White man in his 50s presented with painful oral erosions that began 2 years before. Medical history included recurrent respiratory infections and diarrhea. A previous intestinal biopsy had shown unspecific findings and absence of apoptosis. He reported a 5-kg weight loss attributed to impaired intake due to oral lesions. He was being evaluated for a mediastinal mass detected during chest radiography performed for pneumonia.
</case_information>

<physical_examination>
Physical examination revealed whitish edematous lacy patches with erosions on the dorsal and lateral aspects of the tongue and buccal mucosa. Skin and nails were not involved.
</physical_examination>

<diagnostic_tests>
- Biopsy & Immunofluorescence:
  - Lingual biopsy showed epidermal acanthosis with scattered necrotic keratinocytes, basal-layer vacuolation, and band-like lymphohistiocytic infiltrate
  - Direct immunofluorescence showed fibrinogen deposition along basement membrane
  - Indirect immunofluorescence testing (using monkey esophagus, salt-skin split, and rat bladder) was negative
  - ELISA and immunoblot analysis negative for multiple antibodies
- Laboratory Tests:
  - Normal liver function
  - Negative for hepatitis B and C
  - Decreased IgG serum level: 465 mg/dL (normal: 750-1600 g/L)
  - Normal IgA and IgM levels
  - Decreased total B lymphocyte count
  - Increased CD8+ T count with inverted CD4:CD8 ratio
- Imaging:
  - Positron emission tomography revealed a 10-cm hypercapturing extrapulmonary mass
  - Surgical excision and histopathology of mediastinal mass showed type AB thymoma, Masaoka-Koga stage 2A
- Images:
  - Figure A: Lacy, white erosive lesions on the tongue
  - Figure B: Epidermis with acanthosis, parakeratosis, necrotic keratinocytes, and dermal band lymphocytic infiltrate
  - Figure C: Fibrogen deposition along the basement membrane
  - Figure D: Hypercapturing, heterogeneous mass in right anterior hemithorax
</diagnostic_tests>

Please analyze the information and provide your diagnosis.

Guidelines:
- Use standard medical terminology
- If multiple diagnoses apply, list each on a separate line
- Be specific (e.g., "Type 2 Diabetes Mellitus" not just "Diabetes")
- If no clear diagnosis can be made, state "Insufficient information"

Output format:
<thinking>Use internal reasoning to decide diagnosis names but do not reveal the reasoning process.</thinking>
<final_diagnosis>
Primary diagnosis name
[Secondary diagnosis name, if applicable]
</final_diagnosis>

Example:
<final_diagnosis>
Acute Appendicitis
</final_diagnosis>
\end{lstlisting}

\subsection*{A.1 Phase 0: Clinical Fact Extraction}
Structured case facts were produced as a compact source-of-truth object.
Representative output:
\begin{lstlisting}[breaklines=true,basicstyle=\ttfamily\small,columns=fullflexible]
clinical_extraction
main_case_text: A White man in his 50s presented with painful oral erosions ...
timeline: Oral erosions began 2 years prior; mediastinal mass detected incidentally during pneumonia evaluation.
demographics: White man in his 50s with recurrent respiratory infections and diarrhea ...
positives (7): includes oral erosions and biopsy/immunofluorescence findings ...
negatives (3): indirect IF negative; ELISA/immunoblot negative; hepatitis B/C negative ...
diagnostics (7): includes biopsy & immunofluorescence, imaging, and laboratory entries ...
other_information (3): includes weight loss and mediastinal-mass context ...
\end{lstlisting}

\subsection*{A.2 Phase 1: Patient-Level RAG Support}
Patient-level retrieval used hybrid search over \texttt{60} papers and
compiled a compact evidence context used only as supportive input for
downstream reasoning.
Relevant titles used in this phase included:
\emph{Paraneoplastic pemphigus presenting with a single oral lesion},
\emph{Two Cases of Atypical Bullous Disease Showing Linear IgG and IgA
Deposition in the Basement Membrane Zone}, and
\emph{A Comparative Study of Direct Immunofluorescence Patterns in
Linear IgA Bullous Dermatosis Versus Dermatitis Herpetiformis}.

\subsection*{A.3 Phase 2: Pattern-Trigger Output}
Runtime local retrieval returned \texttt{10} reusable pattern triggers
to guide candidate generation (heuristic scaffolding only). These
retrieved triggers are injected into the candidate-generation prompt as
few-shot guidance.
Example triggers from this run included:
\begin{lstlisting}[breaklines=true,basicstyle=\ttfamily\small,columns=fullflexible]
- Painful swelling of proximal nail folds + lichenoid plaques in oral mucosa + positive anti-hepatitis C virus antibody -> include bullous lichen planus of the nails; confirm biopsy showing hyperkeratosis and dense bandlike lymphohistiocytic infiltrate
- Tense bullae and erosions with progressive tetraparesis + immunofluorescence showing linear IgG deposition -> include overlap of bullous pemphigoid, anti-p200 pemphigoid, and anti-laminin-332 mucous membrane pemphigoid; confirm histopathology and immunological studies
- Pruritic blisters and bullae on hands and feet + elevated white blood cell count + positive enterovirus PCR from skin lesions -> include atypical hand-foot-mouth disease; confirm punch biopsy and exclude other infections with cultures and serologies
\end{lstlisting}

\subsection*{A.4 Phase 3: Candidate Generation Output}
The model produced \texttt{4} scored candidates:
\texttt{Good Syndrome (9/10)},
\texttt{Linear IgA Bullous Dermatosis (7/10)},
\texttt{Mucous Membrane Pemphigoid (6/10)}, and
\texttt{Type AB Thymoma (5/10)}.
Representative output:
\begin{lstlisting}[breaklines=true,basicstyle=\ttfamily\small,columns=fullflexible]
batch_candidate_generation
candidate_1: Good Syndrome | score=9 | reason: Hypogammaglobulinemia, recurrent infections, thymoma, and oral erosions with fibrinogen deposition on immunofluorescence
candidate_2: Linear IgA Bullous Dermatosis | score=7 | reason: Fibrinogen deposition at basement membrane, though lacks classic IgA staining and blistering morphology
candidate_3: Mucous Membrane Pemphigoid | score=6 | reason: Oral erosions and lymphohistiocytic infiltrate, but no IgG/C3 deposition noted
candidate_4: Type AB Thymoma | score=5 | reason: Confirmed by histopathology and imaging, but secondary to underlying syndrome
\end{lstlisting}

\subsection*{A.5 Phase 4: Differential Analysis Output}
A shared micro-RAG retrieval over \texttt{50} papers (\texttt{10}
selected) supported a comparative evidence memo over all candidates.
\begin{lstlisting}[breaklines=true,basicstyle=\ttfamily\small,columns=fullflexible]
query: Direct immunofluorescence fibrinogen deposition interface dermatitis hypogammaglobulinemia thymoma

differential analysis:
The patient presents with a complex constellation of symptoms and
findings suggestive of an underlying immune-mediated disorder associated
with a thymoma. Key features include chronic oral erosions with
characteristic histopathological changes---epidermal acanthosis,
necrotic keratinocytes, basal-layer vacuolation, and a band-like
lymphohistiocytic infiltrate---consistent with autoimmune blistering
disease such as paraneoplastic pemphigus.

Direct immunofluorescence demonstrates fibrinogen deposition along the
basement membrane zone, supporting an autoimmune mechanism targeting
epithelial structures.

... (remaining memo text continues in the original log output)

Sample related article titles from this differential phase:
- Radiologic changes in the thymoma-hypogammaglobulinemia syndrome.
- B-cell lymphopenia and hypogammaglobulinemia in thymoma patients
- Thymoma Associated with Hypogammaglobulinemia (Good's Syndrome): Report of a Case
- Thymoma with chronic diarrhea
\end{lstlisting}

\subsection*{A.6 Phase 5: Final Decision Output}
Final structured output selected
\texttt{selected\_diagnosis="Good Syndrome"} with
\texttt{confidence=9/10}.
Representative output:
\begin{lstlisting}[breaklines=true,basicstyle=\ttfamily\small,columns=fullflexible]
final_decision
selected_diagnosis: Good Syndrome
confidence: 9/10
reason: Clinical features align closely with Good Syndrome: thymoma, hypogammaglobulinemia (IgG 465 mg/dL), recurrent infections, and immune dysregulation (inverted CD4:CD8 ratio, elevated CD8+ T cells). Oral erosions with fibrinogen deposition on immunofluorescence support an autoimmune mechanism secondary to thymoma. Other candidates lack critical corroborative evidence: Linear IgA Bullous Dermatosis fails to demonstrate IgA deposition, Mucous Membrane Pemphigoid lacks IgG/C3 deposition, and Type AB Thymoma alone does not explain systemic immune dysfunction. Verbatim support: 'Weight loss due to impaired intake from oral lesions' (linked to immunodeficiency) and 'Direct immunofluorescence showing fibrinogen deposition along basement membrane' (supports autoimmune etiology).
\end{lstlisting}

\subsection*{A.7 Run Summary}
\texttt{total\_llm\_calls=10}, \texttt{unique\_candidates=4}.
The final diagnosis was selected from explicit candidates using
quote-anchored evidence constraints.

\bibliography{references}

\end{document}